\documentclass[3p,times,12pt]{elsarticle}
\usepackage{etoolbox}
\makeatletter
\patchcmd{\ps@pprintTitle}{\footnotesize\itshape
       Preprint submitted to \ifx\@journal\@empty Elsevier
       \else\@journal\fi\hfill\today}{\relax}{}{}
\makeatother

\usepackage{amssymb}
\usepackage{float}
\usepackage{lineno}

\usepackage[linesnumbered,ruled,vlined]{algorithm2e}
\usepackage{arevmath}     
\usepackage[noend]{algpseudocode}
\usepackage{array}
\usepackage{cellspace}
\usepackage{multirow}
\usepackage[hypcap=false]{caption}
\usepackage{lineno}
\usepackage{footnote}
\usepackage{xcolor}
\usepackage{amssymb}
\usepackage{mwe}
\usepackage{graphicx}
\usepackage[figuresright]{rotating}

\newcolumntype{?}{!{\vrule width 1.5pt}}
\newcolumntype{M}[1]{>{\raggedright}m{#1}}

\SetKwInput{KwInput}{Input}                
\SetKwInput{KwOutput}{Output} 

\usepackage{amsmath,amssymb,amsfonts}
\usepackage{graphicx}
\usepackage{textcomp}

\begin{document}

\begin{frontmatter}


\title{ A Hybrid APM-CPGSO Approach for Constraint SatisfactionProblem Solving: Application to Remote Sensing }


\author[a,b]{Zouhayra Ayadi} 
\author[a,c]{Wadii Boulila}
\author[a]{Imed Riadh Farah }

\address[a]{ RIADI Laboratory, National School of Computer Science, University of Manouba, Manouba 2010, Tunisia}
\address[b]{ICube Laboratory, University of Strasbourg, Strasbourg 67412, France}
\address[c]{IS Department, College of Computer Science and Engineering, Taibah University, Medina 42353, Saudi Arabia}

\begin{abstract}
Constraint satisfaction problem (CSP) has been actively used for modeling and solving a wide range of complex real-world problems. However, it has been proven that developing efficient methods for solving CSP, especially for large problems, is very difficult and challenging. Existing complete methods for problem-solving are in most cases unsuitable. Therefore, proposing hybrid CSP-based methods for problem-solving has been of increasing interest in the last decades. This paper aims at proposing a novel approach that combines incomplete and complete CSP methods for problem-solving. The proposed approach takes advantage of the group search algorithm (GSO) and the constraint propagation (CP) methods to solve problems related to the remote sensing field. To the best of our knowledge, this paper represents the first study that proposes a hybridization between an improved version of GSO and CP in the resolution of complex constraint-based problems. Experiments have been conducted for the resolution of object recognition problems in satellite images. Results show good performances in terms of convergence and running time of the proposed CSP-based method compared to existing state-of-the-art methods.
\end{abstract}


\end{frontmatter}


\section{Introduction}
\label{main}
The constraint satisfaction problem (CSP) \cite{schiex2000reseaux} is essentially a constraint-based paradigm that provides a framework for modeling and solving complex real-world problems. It is an active and growing field \cite{tsang2014foundations} whose objective is to assign values to the various variables constituting the problem \cite{freuder2006constraint}, intending to find one or more solutions that satisfy all or most of the constraints.\\
In the literature, several methods have been proposed to solve CSPs, which can be classified into two main groups: (i) \textit{Complete resolution methods}  \cite{salido} : they include backtracking (BT) \cite{Golomb}, prospective methods \cite{prosser} (such as constraint propagation and filtering algorithm), retrospective methods (Conflict directed backjumping \cite{chen}, dynamic BT \cite{verf} and Nogood-Recording (NR) algorithm\cite{schiex}), selective heuristic \cite{potts} and more others. These are declarative methods that model complex problems in an expressive manner and systematically explore the entire research space to identify the set of solutions or prove the non-existence of solutions. However, they are time-consuming. For this reason, they are unsuitable, especially for large problems. And, (ii) \textit{incomplete resolution methods} \cite{salido}: They are based on local search methods \cite{rossi} and metaheuristics \cite{dreo}, which represent an alternative to complete methods. They solve complex problems with a large search space within a reasonable time. However, they are limited to exploring only a few search space zones.

Both families of resolution methods are very complementary \cite{lamb}. The performance of the complete methods is hampered by the enormous computation time consumed \cite{salido}. Whereas, incomplete methods allow gains in the solving time at the expense of the efficiency of solutions \cite{rossi}. This has prompted various research communities to pay more attention to hybridizing the two families in different ways to reap the advantages of each method and to overcome their drawbacks. In recent decades, several incomplete methods have been proposed in the context of solving CSPs. Among the most successful methods, we can mention the Genetic Algorithm (GA) \cite{sastry}, Ant Colony Optimization (ACO) \cite{sast}, and Particle Search Optimization (PSO) \cite{kennedy1995particle}. More recently, Focus Group Optimization (FGO) \cite{Bi2020} and Group Search Optimization (GSO) \cite{he2009group} have been proposed.
GSO has proven its performance in various applications, such as, medicine \cite{rafi2016optimal}, classification \cite{Ra2017feat}, engineering \cite{Ah2015}, scheduling \cite{Gu2016}, and more. However, it has rarely been used to solve constraint-based problems. Practically, no work has focused on integrating GSO into a constraint programming language (or hybridization).\\
In this context, we aim to combine the exact CP method and an improved version of GSO algorithm. The first one allows to reduce the search space and consequently to face the weak point of time-consuming. Though, instead of guiding the search for solutions using the standard BT method, it will be based on the GSO metaheuristic.

This paper presents, to the best of our knowledge, the first paper proposing a hybridization between GSO and complete methods to solve complex constraint-based problems. It is applied to the resolution of object recognition problems in satellite images.
The rest of this paper is organized as follows. Section 2, briefly, recalls the definitions of each of the concepts CSP, CP, and GSO. The third section presents the related work that has been proposed in the literature. The fourth section describes and details the proposed hybrid CSP resolution algorithm. The fifth section presents experimental results. Finally, conclusion and directions for future work are presented in Section 6.

\vspace{-0.3cm}
\section{Related work and contributions}

Over the last decades, a number of research works have been devoted to the use of the GSO algorithm to solve complex problems, either by improving it, extending it, or modifying it. For example, Balakrishnan and Karthikeyan \cite{balakrishnan2019microarray} use a refined GSO algorithm in the medical domain to find the best multi-category classifications of cancer genetic data sets. In \cite{wang2015improved}, the author presents a simplified GSO algorithm (SGSO) that has been tested on benchmark functions. Chen et al. \cite{chen2016variant} propose a new variant of GSO (VGSO) for solving complex optimization problems that have been evaluated on benchmark functions. In \cite{Ra2017feat}, GSO was used to classify plant neural systems based on leaves.

While these works have shown excellent performance for solving complex unconstrained problems, they have rarely been used for solving constrained problems. In 2011, Shen et al. \cite{shen2011group} started to apply the basic GSO algorithm in solving optimization problems with constraints. Later, in 2013, Wang and Zhang \cite{wang2013modified} proposed a modified version of GSO, incorporating three strategies, (i) a pick flight strategy, (ii) a dimension-by-dimension search, and (iii) a mutation strategy. This algorithm was tested on thirteen benchmark functions based on the known constraints. In \cite{alipour2015improved}, Alipour et al. proposed an improved version of GSO applied to electrical system problems. The objective is to find an optimal solution to the overcurrent relay coordination problem which represents very constrained optimization problems. More recently, in \cite{lakshmanna2016constraint}, authors used GSO in a specific module of the constraint-based DNA sequence extraction process.

Solving complex constraint-based problems is an active and current research area. However, in practice, managing constraints is tricky. Therefore, the solving phase of these problems is a key phase in the CSP treatment process.  Hence, it is essential to improve methods of solving CSPs and to pay attention to the development of hybrid resolution methods. In particular, various studies combining the complete CP method with metaheuristics (such as GA \cite{mirjalili2019genetic},
PSO 
\cite{kennedy1995particle}, 
and ACO \cite{dorigo1999ant, dorigo2019ant}) have been proposed to solve constraint-based problems. In the following, relevant papers that have been conducted in this context are presented. In \cite{di2013hybrid} and \cite{groleaz2020solving} CP was hybridized to ACO. In the first work, the objective was to solve a real-world bicycle sharing problem. While the second study aims to solve one of the variants of the food industry's order scheduling problems.
Fallahi et al. \cite{fallahi2020tabu} proposed an approach based on the hybridization between CP and tabu search to solve complex constraint-based scheduling and routing problems whose objective is to minimize the distance traveled to serve customers. 
In \cite{aastrand2020underground}, the authors proposed a hybrid method based on CP and Large Neighborhood Search (LNS) to solve the automatic planning problems of underground mines.
During the last decades, some works have started to extend the GSO algorithm by combining it with different algorithms. In 2014, Wang and Li \cite{wang2015improved} proposed a hybrid algorithm, named Group Search-artificial Fish Swarm Algorithm GS-AFSA based on GSO and Artificial Fish Algorithm (AFSA), which was applied to three real examples from the field of structural engineering. Later, in 2017, Parthasarathy and Venkateswaran \cite{parthasarathy2017deadline} proposed a method hybridizing GSO and Center-Based Genetic Algorithm (CBGA) for task planning. In the same year, the authors in \cite{lakshmanna2018mining} optimized DNA sequence extraction by proposing a method combining GSO with the Firefly Algorithm (FA). In 2018, Nanivadekar and Kalekar \cite{nanivadekar2018hybrid} combine GSO with Grey Wolf Optimization (GWO) to allocate resources for cognitive radio systems.
In \cite{xue2019estimation}, GSO is combined with Singular Value Decomposition (SVD) to estimate low-frequency oscillation parameters in electrical networks.
Table \ref{comp} summarizes and compares these different works.
\begin{table}[!ht]
\small
\caption{Comparison of related work combining GSO with other methods}
\label{comp}       
\begin{tabular*}{\hsize}{@{\extracolsep{\fill}}llcll@{}}
\hline
\textbf{Reference}	&  \textbf{Methods}& \textbf{Constraints} & \textbf{Application field} & \textbf{Combination}\\	
\hline
Wang and Li \cite{wang2015improved}&AFSA and GSO & - & Structural engineering & Integrative\\
\hline
Parthasarathy &CBGA and GSO & + & Ask scheduling in a cloud & Integrative\\
and al. \cite{parthasarathy2017deadline}&&&&\\
\hline
Lakshmanna  & GWO and GSO & + & Optimization of the DNA& Homogeneous 
 \\
  and Khare \cite{lakshmanna2018mining} & &  &  sequence extraction process &  collaborative
 \\
\hline
Naniva and al. \cite{nanivadekar2018hybrid} &FA and GSO
 & - & Resource allocation of the  & Homogeneous \\
  &
 &  & cognitive radio system & collaborative\\
\hline
 Xue and al. \cite{xue2019estimation} & SVD and GSO
 & - & Identification and determi-  &Heterogeneous\\
  &
 &  & nation of mode parameters   & collaborative\\
&
 &  & in an oscillation event  &  \\
\hline
\end{tabular*}
\end{table}

However, to the best of our knowledge, there is no work that has combined the GSO algorithm with complete methods. This study proposes to test the efficiency of this combination. In the following section, we highlight the motivations that encourage us to combine the two methods CP and an improved version of GSO, as well as the contributions of this paper.
Various motivations encouraged us to propose a hybrid method for CSP solving: 
(1) GSO has been applied in several domains such as networking, energy allocation, scheduling and planning problems, and others, and it has always provided satisfactory results. However, it has never been performed for solving CSPs in the field of remote sensing. (2) Metaheuristics are classified into two sub-families: PSOM and LSOM. The GSO algorithm is characterized by the fact that it is based on both techniques \cite{abualigah2020group} at the same time in its execution process. This intelligent search allows it to find the best solutions. (3) Moreover, it does not require numerous parameters to be set at the beginning. This makes it simple to run and easy to implement. (4) GSO has been combined with other methods but has never been combined with complete methods, although these latter have been combined with several other metaheuristics and have been shown to give very high-performance results. (5) Complete methods such as filtering (or CP) are often unable to solve large complex real-world problems in a reasonable time.

All these reasons have motivated us to hybridize the two methods GSO and CP to take advantage of the strengths of each of them. The proposed method is generic, but we choose to apply it in the field of remote sensing, which is a domain rarely applied for the resolution of CSPs. This constitutes another added value of the current study.
\vspace{-0.55cm}
\section{Background}
\subsection{Constraint satisfaction problem (CSP)}
CSP expresses complex problems in terms of constraints linking their variables and then solves these problems using a set of resolution methods \cite{tsang2014foundations}.
Real complex problems use data that are imperfect in nature \cite{W2009,W2014,W2015,F2017}. Dealing with this kind of problem  has become a research area of great importance that has attracted the attention of most scientific researchers \cite{W2018} and has given rise to new variants of CSP, including, the Fuzzy CSP (FCSP),probabilistic constraint satisfaction problem (VPCP), stochastic constraint satisfaction problem (SCSP). \\
A classical CSP is defined by a triplet $N = (X, D, C)$
\cite{dechter1992constraint} wherein:
\begin{itemize}
\item $ X = \{x_{1},\dots,x_{n}\}$ is the set of variables defined respectively on domains $D (x_{1}),\dots,$ $D(x_{n})$ and subjected to a set of constraints $C$.
\item $D$ represents the set of domains associated with the variables of $X$. Each domain $D_{i}$ represents the values that can be affected to $x_{i}$ with $i \in [1 ..n]$.
\item $C$ is the set of constraints. A constraint is a relationship among a set of variables that each has a domain of possible values. A constraint $c \in C$ is defined by a pair $c= (V(c),R(c))$, where $V(c)=\{x_{i_{1}},\dots,x_{i_{k}}\}\subseteq X$ ($k$ corresponds to the arity of the constraint) is the set of variables that are involved in $c$ and $R(c)$ is a relation defined on the Cartesian product of the domains associated with the variables $V(C)$ $P=D (x_{i_{1}}) \times \dots \times D (x_{i_{k}})$ and which represents the constraint $c$. Otherwise, $R(c)$ represents a subset of values $R(c) \subseteq D^{V(c)} $.
\end{itemize}
The solution's quality often depends on the solving task. It is therefore important to focus on the process of solving complex real-world problems and especially on the selected resolution methods .

\subsection{Group Search Optimisation (GSO)}
Metaheuristics, as one of the sub-families of incomplete CSP resolution methods.

In particular, GSO is, one of the nature-inspired metaheuristics, proposed by \cite{he2006novel,he2009group}. It is based, mainly, on the producer-scrounger model and the scanning technique inspired by the theory of animals' group living and their social foraging behavior. 
It allows, through a well-determined strategy and a set of operators, to generate the best solutions. This algorithm is mainly based on three types of operators: the \textit{producer} is the member with the best value of F and who executes a production strategy to find the most promising opportunities. The \textit{scroungers} represent a group of members, other than the producer, who uses a scrounging strategy to join the resources discovered by the producer. And, the \textit{rangers} represent the rest of the members who use a search strategy by moving randomly in the search space. This operator has been added to avoid falling into local minima.
 
Members, initially generated, are iteratively improved, based on an evaluation function, and only the good solutions are extracted. 

\vspace{-0.3cm}
\section{The hybrid proposed CSP algorithm}
The proposed approach is essentially based on two main modules: (i) the CP or filtering module and (ii) the GSO search module. In this section, we describe the main steps of the proposed hybrid APM-CPGSO approach.
\begin{enumerate}
\item \textbf{Parameter initialization and CSP modeling}\\
The proposed method starts with two preprocessing steps: \\
 $\bullet$ The generation of the CSP consisting of defining the various variables of the problem, their domains of possible values, and the definition of the various constraints.\\
$\bullet$  The initialization of parameters of the proposed GSO algorithm. Values are set based on previous works \cite{he2009group, kennedy1995particle, Bi2020} (default setting parameters).
\item \textbf{Generation of the initial population}\\
The second step  consists of randomly generate a population $G$: a group of members, where each member is defined by a head angle $\theta_{i}^{k}=\theta_{i1}^{k},\dots,\theta_{i(n-1)}^{k}$ and has a current position $Y_{i}^{k} \in R^{n}$. At each iteration, the member follows a search direction $G_{i}^{k}(\theta_{i}^{k})=(g_{i1}^{k},\dots,g_{in}^{k})$ calculated based on the search angle $\theta_{i}^{k}$ via a polar coordinate transformation using following equations:
\begin{equation}
g_{i1}^{k}= \prod_{q=1}^{n-1} Cos(\phi_{iq}^{k})
\end{equation}
\begin{equation}
g_{ij}^{k}= Sin(\phi_{i(j-1)}^{k}).\prod_{q=1}^{n-1} Cos(\phi_{iq}^{k})
\end{equation}
\begin{equation}
g_{in}^{k}= Sin(\phi_{i(n-1)}^{k})
\end{equation}
\begin{figure}[ht!]\vspace*{4pt}
\centerline{\includegraphics[height=9.6cm,width=0.8\textwidth]{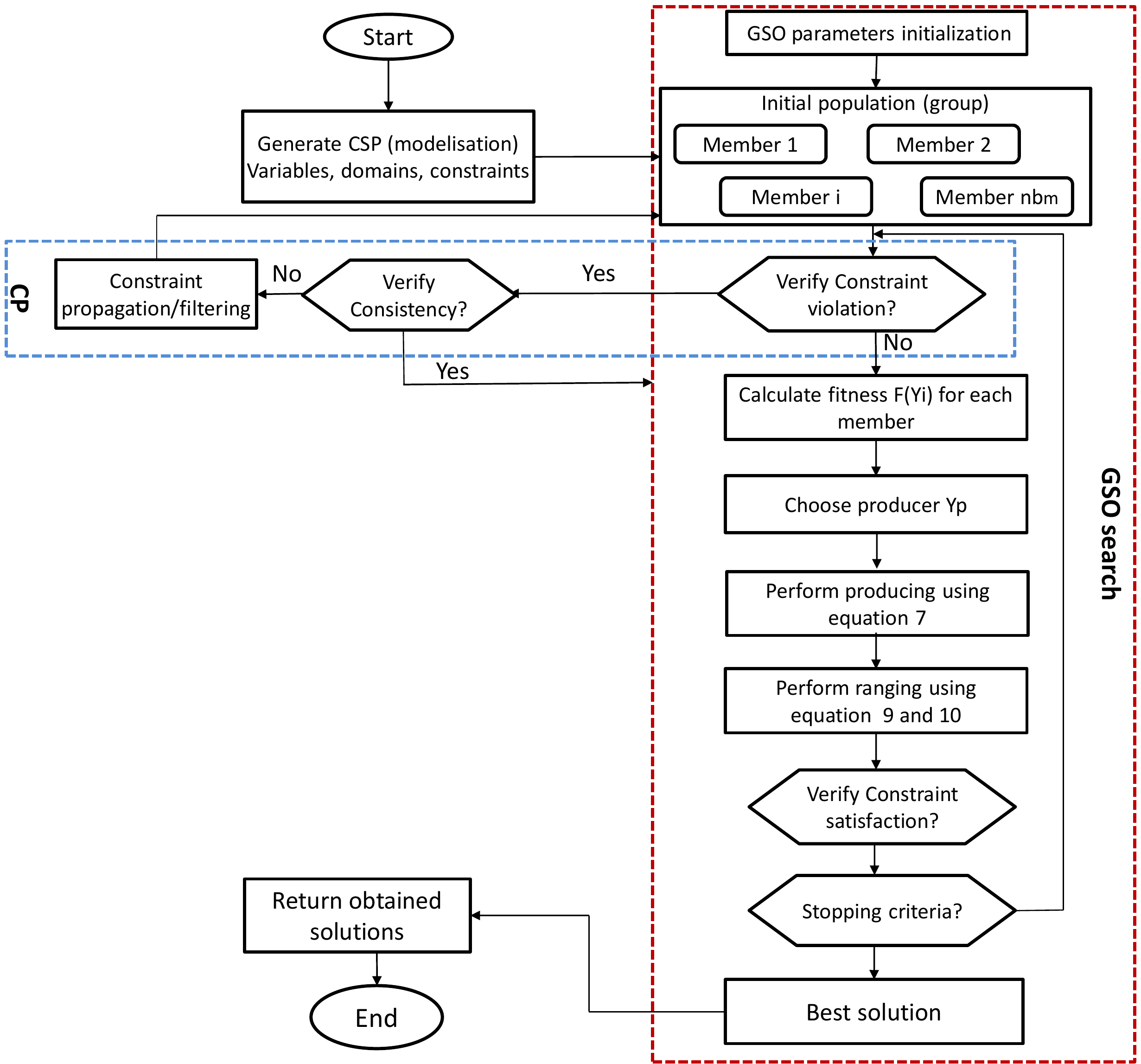}}
\centering
\caption{Flowchart of the proposed approach}
\label{Flow}
\end{figure}
\\
For a CSP, the individuals constituting the group $G$ represent parts of the search space. Each member of the population $m \in G$ represents a subset of the domain, i.e. a complete assignment (or a potential solution). Each variable will, therefore, be associated with a subset $d_{i}$ of its domain $D_{i}$,
$m= {(d_{1},...,d_{n})}| \forall 1 \leq i \leq n , d_{i} \subseteq D_{i}$\\
The initial population is generated after the CSP is modeled, i.e.  after the variables are generated, the domains defined and the constraints listed.  During the initialization of this population(the assignment of values to the variables), the arc consistency is executed in order to remove some values from the domains of the related variables.  However, since the generation of the population is done randomly, it is composed of feasible and non-feasible members.  In this way the diversity of the initial group is maintained and the convergence of the GSO search module is improved and accelerated.  For this reason, the solution search module is combined with a filtering module which will be described later

\item \textbf{Computation of the evaluation function}\\
To evaluate each member of the initially generated group, an evaluation function must be calculated. This function $F$, also called evaluation function, often depends on the problem to be treated. Since CSPs are modeled as a network of constraints, it is often difficult to find solutions that satisfy all the constraints.\\
As mentioned in the previous section, solving a CSP comes down to finding the members (the tuple of values assigned to all the variables) that satisfy all the constraints or as many constraints as possible. Otherwise, it is a matter of finding the set of solutions $S$ defined by the complete instantiations satisfying the constraints $C_{j}$ with $j \in [1..m]$: $S \equiv {(x_{1}= v_{1}),\dots,( x_{n}= v_{n})}$ where $x_{i}= v_{i}$ with $v_{i} \in D_{i}$ means that each variable $x_{i} $ is assigned to a value in its domain.
The evaluation function must estimate the measure to which the constraints are violated and is defined as the total number of violated constraints (which must be minimized) or the total number of satisfied constraints (which must be maximized) for each member or potential solution.
\item \textbf{Constraint propagation (CP) module}\\
When generating the initial population, each member forms a complete instantiation where each variable is associated with a value. However, these generated assignments may be inconsistent. Yet, the goal of members is to provide quality and valid value tuples. \\
Constraint propagation or filtering \cite{dechter2003constraint} is one of the most widely used methods of complete CSP resolution. Each time a value is assigned to a variable, this method anticipates the effect of the current partial assignment on the rest of the unassigned variables, filters their domains from inconsistent values, and improves the search space exploration phase. The propagation process stops and returns an error message if the domain of a variable becomes empty.\\
The integration of the CP module will allow acting on the initial population by filtering certain inconsistent values of the related variables’ domains. This will provide reduced search areas  and will make the GSO more optimal. As a result, the GSO will explore the search space more accurately and quickly.\\
The process is illustrated in Fig. \ref{Flow}. By assigning values to the variables, the adequacy of the constraint’s satisfaction can be checked. If they are not violated, the GSO search module is started. In case of non verification of the consistency, the assignment is given to the initial population. In case the consistency is verified, the search module is started.\\
Various filtering algorithms have been proposed such as Forward Checking (FC), node consistency, arc consistency, etc. In this study, arc consistency is used because it provides better results, especially for large problems. In addition, to obtain global consistency, the classical search algorithm is replaced with an incomplete method, the improved GSO that will be detailed in the next section.

\item \textbf{GSO Search module}\\
This module is used to guide the search for the best solutions. 
\begin{itemize}
\item \textit{Choose the producer and perform producing:}
the member with the best evaluation function value  is chosen as the producer. It executes the producing strategy by analyzing the promising search space around its current position, according to three points: zero degree, right, and left
\begin{equation}
Y_{Z}= Y_{q}^{k}+ r_{l} l_{max} G_{q}^{k}(\theta^{k})
\end{equation}
\begin{equation}
Y_{R}= Y_{q}^{k}+ r_{l}l_{max} G_{q}^{k}(\theta^{k}+\frac{r_{2} \phi_{max}}{2})
\end{equation}
\begin{equation}
Y_{L}= Y_{q}^{k}+ r_{l}l_{max} G_{q}^{k}(\theta^{k}-\frac{r_{2} \phi_{max}}{2})
\end{equation}
Instead of following the standard three-point production search method which is too computationally intensive, we improve GSO by using Adaptive Probability Mutation. This method is based on the polynomial probability distribution:
\begin{equation}
mx_{Pj}= x_{pj}^{k}+ (x_{ij}^{U}-x_{ij}^{L}) \times \delta
\end{equation}
where $x_{ij}^{U}$ is the upper bound and $x_{pj}^{L}$ the lower bound of $x_{pj}$. 
In the  proposed method, APM is used in local search area with mutation probability:
\begin{equation}
    P_{m}= \frac{1}{d}+ \frac{k}{k_{max}}(1-\frac{1}{d})
\end{equation}
where $d$ is the dimension of problem, $k$ current iteration and $x_{max}$ maximum iteration number.
\item \textit{Perform scrounging:}
40\% of members are selected as scroungers. They move towards the producer according to the following equation:
\begin{equation}
Y_{i}^{k+1}= Y_{i}^{k}+ r_{3} \circ (Y_{d}^{k} - Y_{i}^{k})
\end{equation}
\item \textit{Perform ranging:}
The remaining members are selected as rangers. They generate a random head angle, choose a random distance, and move randomly according to the following equations:
\begin{equation}
\theta_{i}^{k+1}= \theta_{i}^{k}+ r_{2} c_{max} 
\end{equation}
\begin{equation}
l_{max}= c r_{l} l_{max} 
\end{equation}
\begin{equation}
Y_{i}^{k+1}= Y_{i}^{k}+ l_{i} G_{i}^{k}(\theta^{k+1})
\end{equation}
\end{itemize}

\item \textbf{Stop criterion}\\
The stopping criterion depends on the objective of the decision-maker (looking for a solution, all solutions, or the optimal solution).
Three cases arise,
\begin{itemize}
    \item $G= \emptyset$: the search space has been covered in its entirety and all solutions have been found. Otherwise, we haven't individual in the cover population $G$.
\item $ nb_{S}= integer$ or $S \notin \emptyset$ In this case, we choose to stop as soon as we found the first solution, the variable $S$ is not empty anymore, or we need a determined number of solutions and in this case, we stop as soon as $nb_{S}$ is equal to the integer initially declared (the number of the desired solution is established).
\item Iteration $I=nb_{max}$: We can stop the execution of the algorithm at a determined number of iterations $nb_{max}$, whatever the number of solutions found (no solution, one or more).
\end{itemize}
In our context, our objective is to find all the solutions.
The diversity of the choices of stopping criteria makes it possible to manage the search between a systematic search covering the entire search space, and an incomplete search covering only a few promising areas. At the end of the execution, we will always have a double result: the set $S$ of the found solutions ($Sol= S$), as well as a set of potential remaining solutions represented by the current covering population G ($Sol= S \cup Sol(G)$).
\end{enumerate}
\begin{algorithm}[ht!]
\footnotesize
\DontPrintSemicolon
  Set value of parameters $\theta_{max}$, $\alpha_{max}$, $l_{max}$ 
  $K=0;$\\
  Generate randomly initial group: initial $Y{i}^{k}$ and $\theta_{i}^{k}$;\\
  Compute the value of f($Y_{i}$) to evaluate each of initial members;\\
  \If{(Violated constraints}
  {R $\leftarrow$ Verif-consistency(members);\\
  \If{R=true}{
  Filtering(members-domains);
  }
}
  
 \While{(The stop criterion are not satisfied)}{
  \For{(For each member in $N_{pop}$)}
{ 
Select the producer $Y_{p}$;\\
r$\leftarrow$[0,1];\\
  \If{(r $\leq$ Pm)}
  {Apply polynomial mutation to $Y_{p}$ using equation 7;\\ 
  \If{F($Y_{i}$) $\leq$ F($Y_{p}$)}{
  $Y_{p} \leftarrow Y_{i}$}
  }
  
 \If{($rand < 0.6$)}
  {
  Scrounger randomly walks to join the producer according to his current position using equation 9;
     }
    \Else
    { 
    Rangers randomly searching dispersed resources strategy using equation 10-12;

        }
        }
        }
         K $\leftarrow$ K+1;\\Return Solution;
         
   \caption{APM-CPGSO}
   \label{TBT}
\end{algorithm}
\vspace{-1cm} \section{Experimental results}
\begin{figure}[ht!]
\centering
\includegraphics[height=4cm,width=0.99\textwidth]{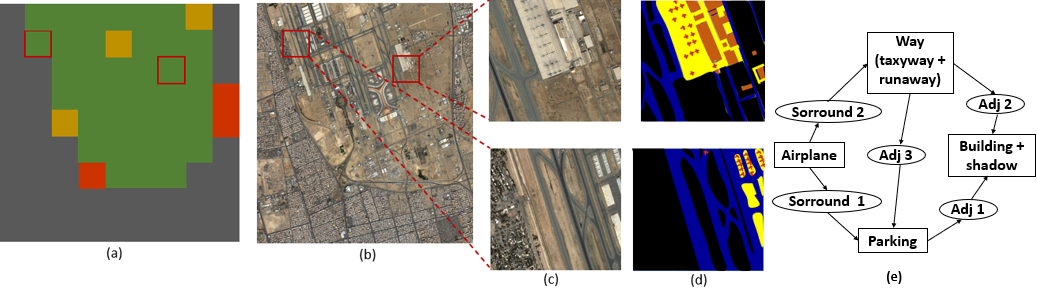}
\caption{(a) Image with detected objects. (b) Original image of a airport. (c) Two original tiles, (d) End result of images in (c) (instantiations). (e) General model of the airport.  }
\label{satt}
\end{figure}

\vspace{-0.3cm}
This section is divided into two parts. The first part consists of applying the proposed approach to solve the object recognition problem in satellite images.
Experiments were performed on an Intel Core i5 computer, 2.3 GHz with 4 GB RAM. The maximum number of iterations of the algorithms was set to 100 executions. The parameters of the different algorithms were fixed by referring to \cite{he2009group, kennedy1995particle, Bi2020}.
We first apply the proposed hybrid method for airport recognition problem on a satellite image produced by the SPOT7 satellite with a spatial resolution of 1.5m. In our case the airport is modeled by a spatial graph (general model of the airport (Fig. \ref{satt} (e))), where vertices represent the simple objects that compose the airport we instantiate in the image (the set of variables X), whilst arcs represent the spatial relationships among the objects, and which are translated into constraints (the set C) to satisfy. Initial domains D of each variable are determined from the image (Fig. \ref{satt} (b)) and correspond to the regions of each object resulting from the segmentation process. The images was acquired on 18-02-2019 and  has been corrected both radiometrically and geometrically. 
It has a spatial resolution 7233$\times$8890 pixels, which was divided into 99 tiles of 805$\times$805 pixels.\\
The proposed approach was applied to all tiles. Fig. \ref{satt} illustrates the application of the proposed approach on the whole image and proves its efficiency. From Fig. \ref{satt} (a), we can note that among all the tiles, 93 were correctly classified: 37 true positives (green tiles) whose defined model (graph model of the airport that we have defined based on simple objects (variables) and the spatial relation between them) was, suitably, detected and 56 true negatives (grey tiles) were suitably undetected. However, only 5 tiles were not correctly classified: 3 false positives (red tiles) where the method identified the model incorrectly and 2 false negatives (yellow tiles) where the method could not detect the model. 
In order to present the obtained results in detail, we illustrate, in Fig. \ref{satt} (c) and (d), two examples of tiles and the obtained instancitations, respectively. The ways (runaways and taxiways) are represented in blue, the buildings in orange, the airplanes in red and the parking areas in yellow.

A set of experiments are conducted to evaluate the performance of the proposed approach and to compare it with state-of-the-art algorithms namely standard GSO, PSO, and FGO. These algorithms are selected based on the following reasons:
\begin{itemize}
\item GSO is selected to highlight the role of the hybridization of GSO with CP. This will confirm our choice of using CP in the proposed approach.
\item FGO is selected since it is a recently proposed nature-inspired incomplete solving algorithm that has shown efficient results in solving CSP.
\item PSO is selected since it has shown that it provides good results compared to the standard GSO algrithm.
\end{itemize}
Figure \ref{conv} depicts the convergence performance of the considered approaches.  We can see that both standard GSO and PSO algorithms start the exploration of the search space very far from the best solutions and converge slowly. FGO and CP-GSO converge, both, more quickly to the optimal values and have the best convergence performances. Additionally, we notice that the proposed method, CP-GSO, starts the search very close to the optimal values and converges in a better time span compared to FGO. This behavior is due to the behavior of CP, which leaves only the feasible potential solutions, and to the improvement in the GSO version. On the contrary, PSO and standard GSO have difficulties to target the feasible areas of the search space, especially since they start with initial populations containing consistent and inconsistent potential solutions, and this explains their low convergence rates.\\
\hspace{-1cm}
    \begin{minipage}{.5\textwidth}
    \centering
 \includegraphics[height=5.5cm,width=\textwidth]{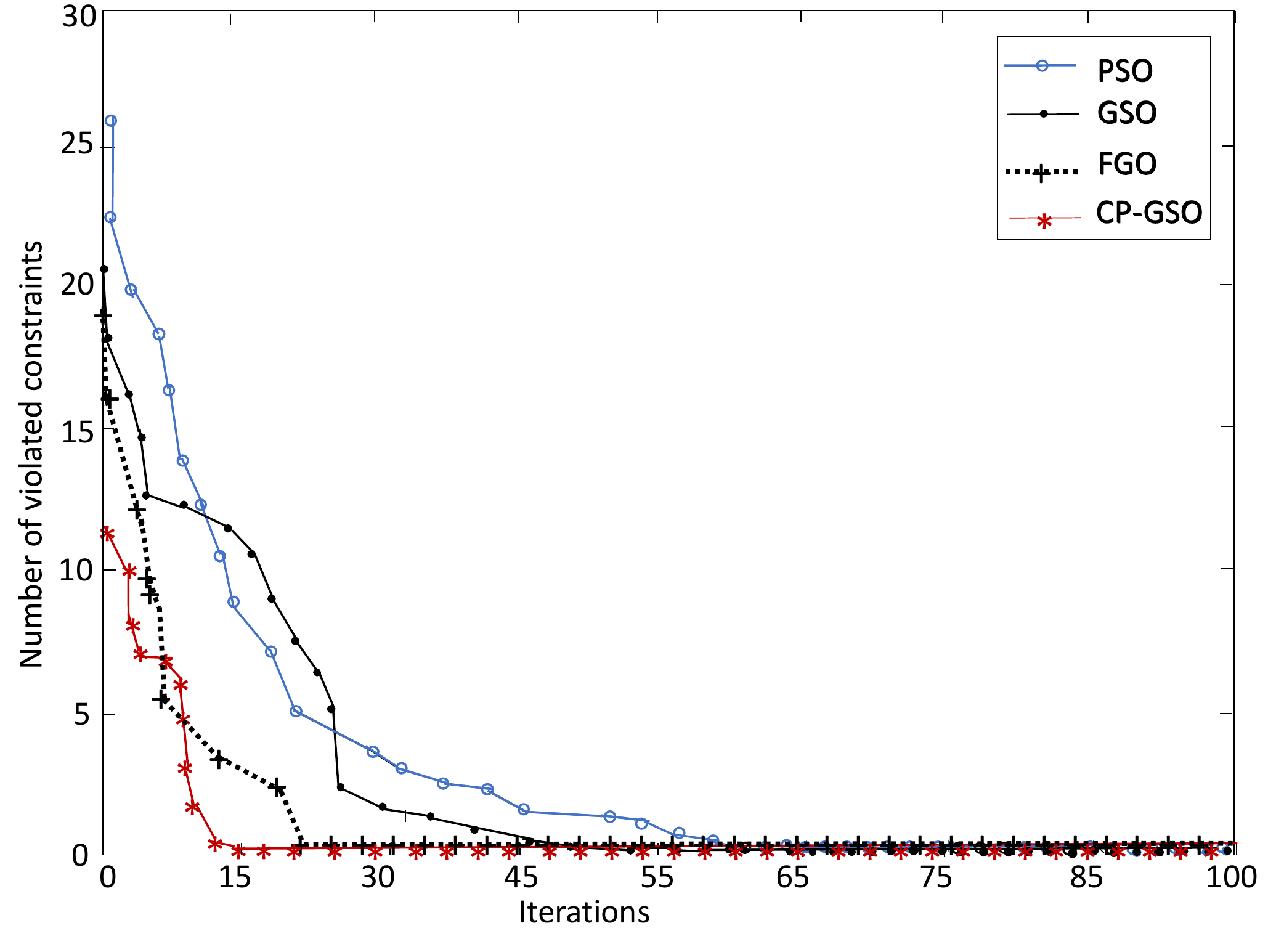}
 \vspace{-0.3cm}
 \captionsetup{width=\linewidth}
\captionof{figure}{ Convergence trend of each algorithm 
}
\label{conv}
\end{minipage}
\hspace{0.3cm}
\begin{minipage}{.45\textwidth}\centering
 \vspace{0.2cm}
    \small
\begin{tabular}{lccc}
\hline
\textbf{Algorithms}	&\textbf{Max}&\textbf{Means}&\textbf{Running time}\\

\hline
APM-CPGSO &11 &1.37 & 23.58  \\
& & &\\
FGO &19 &1.83 & 39.25 \\
& &&\\
Standard GSO& 21 &3.44 & 55.23 \\
& &&\\
PSO & 26&2.99 & 67.16\\
\hline
\end{tabular}
\captionsetup{width=\linewidth}
\captionof{table}{Comparison of experimental results achieved by PSO, standard GSO, FGO, and the proposed method}
\label{cc}
\end{minipage}
     \vspace{0.3cm}
     \\
To further evaluate the efficiency of the proposed approach, we compare the running time of the considered search algorithms  as depicted in Table \ref{cc}. To further evaluate the effectiveness of the proposed approach, we compare the results obtained by the considered search algorithms as depicted in Table 2.  We can see that the computation time and the max value of the evaluation function and the average of APM-CPGSO are more important compared to FGO, standard GSO, and PSO. It has the smallest values of MAX, means, and execution time (11, 1.37,and 23.58 min respectively).

We can, clearly, note that the proposed approach outperforms the other methods. Therefore we can conclude that hybridization has resulted in a better performing algorithm.
 \vspace{-0.3cm}
\section{Conclusion and futur work}
In this paper, a novel hybrid algorithm combining two different families of methods for solving CSPs is presented. The proposed algorithm takes advantages of constraint propagation to reduce the initial population and leave only consistent individuals, and the GSO metaheuristic to guide the search for better solutions.\\
Experiments are conducted to evaluate the performance of the proposed approach for the problem of object recognition from high spatial resolution satellite images. Comparison with state-of-the-art methods shows good results of the proposed algorithm in terms of convergence and running time.\\
The resolution of problems in the context of uncertainty is a major challenge of constraint programming. In the near future, we plan to adapt and implement APM-CPGSO to solve the variants of CSP, including FGSO.


\vspace{-0.3cm}
\begin{thebibliography}{}
\bibitem{Bi2020}
Bidar, M., Mouhoub, M., \& Sadaoui, S. (2020). Discrete FGO Algorithm for Solving CSP. In ICAART (2) (pp. 322-330).
\bibitem{Da2018}
N. Dali, \& S. Bouamama, (2018), "New parallel Genetic Algorithms on GPU for solving Max-CSPs" {\it In 2018 IEEE 14th International Conference on Intelligent Computer Communication and Processing (ICCP)} IEEE, 119-126. 
\bibitem{Gu2021}
Guan, B., Zhao, Y., \& Li, Y. (2021). An improved ACO with an automatic updating mechanism for constraint satisfaction problems. Expert Systems with Applications, 164, 114021.
\bibitem{balakrishnan2019microarray}
R. Balakrishnan and  T.  Karthikeyan, (2019) “Microarray  gene  expression  and multiclass  cancer  classification  using  extreme  learning  machine  (elm) with refined group search optimizer (rgso)", {\it Int Sci J Sci Eng Technol}, {\bf 18}.
\bibitem{wang2015improved}
Y.  Wang  and  L.  Li, (2015) “An  improved  intelligent  algorithm  based  on  the group search algorithm and the artificial fish swarm algorithm", {\it International Journal of Optimisation in Civil Engineering}.
\bibitem{chen2016variant}
J.-J. Chen, T. Ji, P. Wu, and M. Li, (2016) “A variant of group search optimizer for global optimization", {\it JCMSE journal }, {\bf 16}, (2), 219–-230.
\bibitem{rafi2016optimal}
D. M. Rafi and C. R. Bharathi, (2016) “Optimal fuzzy min-max neural network (fmmnn)  for  medical  data  classification using  modified  group  search optimizer  algorithm",{\it International  Journal  of  Intelligent  Engineering and Systems}, {\bf 9}, (3), 1-–10.
\bibitem{shen2011group}
 H. Shen, Y. Zhu, W. Zou, and Z. Zhu,(2011) “Group search optimizer algorithmfor  constrained  optimization", {\it  in International Workshop on Computer Science for Environmental Engineering and Eco Informatics}, Springer, 48–-53.
\bibitem{wang2013modified}
 L.   Wang   and   Y.   Zhong, (2013) “A  modified   group   search   optimiser for constrained optimisation problems,{\it International Journal of Modelling, Identification and Control}, {\bf 18}, (3), 276-–283.
\bibitem{alipour2015improved}
 M.  Alipour, S.  Teimourzadeh,  and  H.  Seyedi, (2015)  “Improved  group  search optimization  algorithm  for  coordination  of  directional  overcurrent  relays",{\it Swarm and Evolutionary Computation}, {\bf 23}, 40-–49.
\bibitem{lakshmanna2016constraint}
K. Lakshmanna and N. Khare, (2016) ``Constraint-based measures for dna sequence mining using group search optimization algorithm",{\it International Journal of Intelligent Engineering and systems}, {\bf 9}, 91–-100.
\bibitem{lakshmanna2018mining}
K. Lakshmanna and N. Khare, (2018) “Mining dna sequence patterns with constraints using hybridization  of  firefly  and  group  search  optimization",{\it Journal  of  Intelligent systems}, {\bf 27}, (3), 349-–362.
\bibitem{mirjalili2019genetic}
 S. Mirjalili, (2019) “Genetic algorithm", {\it in Evolutionary algorithms and neural networks } Springer, 43-–55.
\bibitem{kennedy1995particle}
 J. Kennedy and R. Eberhart, (1995) “Particle swarm optimization", {\it ininternational  conference  on  neural  networks} IEEE, {\bf 4} 1942–-1948.
\bibitem{dorigo1999ant}
  M.  Dorigo  and  G.  Di  Caro, (1999) “Ant  colony  optimization:  a  new  meta-heuristic”,  {\it in Proceedings   of   the   1999   congress   on   evolutionary computation-CEC99  (Cat.  No.  99TH8406)} IEEE, {\bf (2)}, 1470–-1477.
\bibitem{dorigo2019ant}
  M. Dorigo and T. St\"utzle, (2019) “Ant colony optimization: overview and recent advances", {\it Handbook of metaheuristics}, 311-–351.
\bibitem{di2013hybrid}
 L. Di Gaspero, A. Rendl, and T. Urli, (2013) “A hybrid aco+ cp for balancing bicycle  sharing  systems", {\it in International  Workshop  on  Hybrid  Meta-heuristics}    Springer, 198-–212.
\bibitem{groleaz2020solving}
L. Groleaz, S. N. Ndiaye, and C. Solnon, (2020) “Solving the group cumulatives cheduling problem with cpo and aco”, {\it in International Conference on Principles  and  Practice  of  Constraint  Programming } Springer, 620-–636.
\bibitem{fallahi2020tabu}
 A.  E.  Fallahi,  E.  Y.  Anass,  and  M.  Cherkaoui, (2020) “Tabu  search  and constraint programming-based approach for a real scheduling and routing  problem”, {\it International  Journal  of  Applied  Management  Science},{\bf 12}, (1), 50-–67.
\bibitem{aastrand2020underground}
M. Astrand,   M.   Johansson,   and   A.   Zanarini, (2020)   “Underground   mine scheduling of mobile machines using constraint programming and large neighborhood search", {\it Computers and Operations Research}, {\bf123}, p 105036.
\bibitem{parthasarathy2017deadline}
 S.  Parthasarathy  and  C.  J.  Venkateswaran, (2017) “Deadline  constrained  tasks cheduling  method  using  a  combination  of  center-based  genetic  algorithm  and  group  search  optimization”, {\it Journal  of  Intelligent  Systems}, {\bf29}, (1), pp. 53–-70.
\bibitem{nanivadekar2018hybrid}
 S. S. Nanivadekar and U. D. Kolekar, (2018) “A hybrid optimization model for resource allocation in of dm-based cognitive radio system”, {\it Evolutionary Intelligence}, 1-–12.
\bibitem{xue2019estimation}
 Z. Xue, Z. Chen, T. Ji, M. Li, \& Q. Wu, (2019) “Estimation of low frequency oscillation  parameters  using  singular  value  decomposition  combined group  search  optimizer”, {\it Electric  Power  Components  and  Systems}, {\bf 47}, (3), 275-–287.
\bibitem{abualigah2020group}
 L. Abualigah, (2020) “Group search optimizer: a nature-inspired meta-heuristic optimization  algorithm  with  its  results,  variants,  and  applications”, {\it Neural Computing and Applications}, 1-–24.
\bibitem{tsang2014foundations}
 E. Tsang, (2014) Foundations of constraint satisfaction: the classic text. {\it BoD–Books on Demand}.
\bibitem{dechter1992constraint}
 R.  Dechter, (1992) “Constraint  networks”, {\it in:Encyclopedia  of  Artifical  Intelligence}.
\bibitem{dechter2003constraint}
R. Dechter, D. Cohenet al., (2003) Constraint processing.   {\it Morgan Kaufmann}.
\bibitem{he2006novel}
S. He, Q. Wu, and J. Saunders, (2006) “A novel group search optimizer inspired by animal behavioural ecology”, {\it in 2006 IEEE international conference on evolutionary computation} IEEE, 1272-–1278.
\bibitem{he2009group}
 S.  He,  Q.  H.  Wu,  \&  J.  Saunders, (2009) “Group  search  optimizer:  an optimization  algorithm  inspired  by  animal  searching  behavior”, {\it IEEE transactions on evolutionary computation}, {\bf 13}, (5), 973-–990.
\bibitem{Ra2017feat}
A.M. Ravishankkar \& P. Amudhavalli, (2017) "Feature Selection using GSO for Plant Leaf Classification", {\it AJIT Journal}, {\bf 16}, 810--815.
\bibitem{Gu2016}
D. Guanlong, Z. Shuning, \& Z. Mei, (2016) "A discrete GSO for blocking flow shop multi-objective scheduling", {\it Advances in Mechanical Engineering}, {\bf 8} (8).
\bibitem{Ah2015}
A. Ahmadi, A. Kaymanesh, A. Heidari, \& V. G. Agelidis (2015) "Comment on ‘Reliability constrained unit commitment with combined hydro and thermal generation embedded using self-learning group search optimizer", {\it Energy}, {\bf 89}, 1103--1105.
\bibitem{freuder2006constraint}
E. C. Freuder, A. K. Mackworth (2006) "Constraint satisfaction:  An emerging paradigm", {\it in:  Foundations of AI} Elsevier, {\bf2}, 13–-27.
\bibitem{schiex2000reseaux}
T. Schiex (2000), "R\'{e}seaux de contrainte"  {\it HDR, INRA}.
\bibitem{Golomb}
S.  W.  Golomb,  L.  D.  Baumert (1965) "Backtrack  programming",{\it   Journal  of  the  ACM (JACM)}, {\bf12} 516-–524
\bibitem{salido}
R. Bart\'{a}k, M. A. Salido, F. Rossi (2010), New trends in CS, planning,and schedulingy",  {\it The Knowledge Engineering Review}, 249–-279.
\bibitem{prosser}
P. Prosser (1993), "Domain filtering can degrade intelligent BT search",  {\it in: IJCAI, Citeseer}, 262–267.
\bibitem{schiex}
 T. Schiex, G. Verfaillie (1994), "Nogood recording for static and dynamic CSP",   {\it International Journal on AI Tools},  187-–207.
 \bibitem{potts}
 S. C. Brailsford, C. N. Potts, B. M. Smith (1999),  "CSP  Algorithms and applications", {\it European Journal of Operational Research}, 557-–581.
 \bibitem{verf}
 G. Verfaillie, T. Schiex (1994), "Dynamic BT for dynamic CSP ", {\it In Proceedings of Workshop on Constraint Satisfaction, Citeseer}, 1-–8.
 \bibitem{chen}
  X. Chen, and V. B. Peter (2001), "CBJ revisited", {\it Journal of Artificial Intelligence Research}, 53--81.
  \bibitem{sastry}
  K. Sastry, D. Goldberg, \& G. Kendall (2005), "Genetic algorithms", {\it In Search methodologies . Springer}, 97--125).
  \bibitem{sast}
  T. Stützle (2009), "Ant colony optimization", {\it In International conference on evolutionary multi-criterion optimization. Springer}.
   \bibitem{dreo}
  J. Dréo, A. Pétrowski, P. Siarry et E. Taillard (2003), "Métaheuristiques pour
l’optimisation difficile",{\it Eyrolles}.
  \bibitem{rossi}
F. Rossi, P. V. Beek, T. Walsh (2006), "Handbook of constraint programming", {\bf35}.
   \bibitem{lamb}
T. Lambert (2006), "Hybridation de méthodes complètes et incomplètes pour la résolution de CSP", {\it Doctoral dissertation, Nantes}.
\bibitem{W2009}
W. Boulila, I. R. Farah, K. S. Ettabaa, B. Solaiman, \& H. B. Ghezala (2009) ”Improving spatiotemporal change detection: A high level fusion
approach for discovering uncertain knowledge from satellite image databases”, {\it In Icdm}, {\bf9}, 222–-227.
\bibitem{W2015}
W. Boulila, Z. Ayadi, \& I. R. Farah (2017), "Sensitivity analysis approach to model epistemic and aleatory imperfection: Application to Land
Cover Change prediction model", {\it Journal of computational science}, {\bf23}, 58–-70.
\bibitem{W2018}
A. Ferchichi, W. Boulila, \& I. R. Farah (2018), "Reducing uncertainties in land cover change models using SA", {\it Knowledge and Information Systems}, {\bf55}, (3), 719--740.

\bibitem{W2014}
W. Boulila, A. Bouatay, \& I. R. Farah (2014), "A Probabilistic Collocation Method for the Imperfection Propagation: Application to Land Cover Change Prediction", {\it J. Multim. Process. Technol.}, {\bf 5}, (1), 12--32.

\bibitem{F2017}
A. Ferchichi, W. Boulila, \& I. R. Farah (2017), "Propagating aleatory and epistemic uncertainty in land cover change prediction process", {\it Ecological informatics}, {\bf 37}, 24--37.

 
 \end{thebibliography}
\end{document}